# Assessment of Spectral based Solutions for the Detection of Floating Marine Debris

1st Muhammad Alì
*Department of Engineering*
*University of Naples Parthenope*
Naples, Italy
muhammad.ali001@studenti.uniparthenope.it

2nd Francesca Razzano
*Department of Engineering*
*University of Naples Parthenope*
Naples, Italy
francesca.razzano002@studenti.uniparthenope.it

3rd Sergio Vitale
*Department of Engineering*
*University of Naples Parthenope*
Naples, Italy
sergio.vitale@uniparthenope.it

4th Giampaolo Ferraioli
*Department of Science and Technology*
*University of Naples Parthenope*
Naples, Italy
giampaolo.ferraioli@uniparthenope.it

5th Vito Pascazio
*Department of Engineering*
*University of Naples Parthenope*
Naples, Italy
vito.pascazio@uniparthenope.it

6th Gilda Schirinzi
*Department of Engineering*
*University of Naples Parthenope*
Naples, Italy
gilda.schirinzi@uniparthenope.it

7th Silvia Ullo
*Department of Engineering*
*University of Sannio*
Benevento, Italy
silvia.ullo@unisannio.it

*Abstract*—Typically, the detection of marine debris relies on in-situ campaigns that are characterized by huge human effort and limited spatial coverage. Following the need of a rapid solution for the detection of floating plastic, methods based on remote sensing data have been proposed recently. Their main limitation is represented by the lack of a general reference for evaluating performance. Recently, the Marine Debris Archive (MARIDA) has been released as a standard dataset to develop and evaluate Machine Learning (ML) algorithms for detection of Marine Plastic Debris. The MARIDA dataset has been created for simplifying the comparison between detection solutions with the aim of stimulating the research in the field of marine environment preservation. In this work, an assessment of spectral based solutions is proposed by evaluating performance on MARIDA dataset. The outcome highlights the need of precise reference for fair evaluation.

## I. INTRODUCTION

Marine plastic pollution is one of the most terrible environmental problems of recent times which affects oceans and channels within the globe. The extensive presence of plastic debris in marine ecosystems has substantial ecological, financial, and health effects. Yearly, a projected 4.4–12.7 million metric tons of plastic are inserted into the oceans. In 2014, a study assessed that nearly 5.25 trillion plastic is suspended on the marine surface. Plastic contributes 60–80% of the total marine debris, which will reach 90–95% in a few areas and remains hanging in the sea for a hundred to thousands of years due to extended duration and toughness [1].

A rapid tool for the detection of marine debris is fundamental for preserving marine ecosystem. Among many, remote sensing (airborne and spaceborne) is representing a valid solution. Specifically, Earth observation data from both public and commercial satellite programs have been utilized to detect and monitor marine debris [2]. Additionally, remote sensing data from manned aircraft, unmanned aerial vehicles (UAVs), bridge-mounted cameras, and underwater cameras have also been employed. Research on plastic detection using airborne data, models and theoretical studies has demonstrated the potential to detect macro plastics in optical data. Satellite remote sensing is the leading technique for collecting high-quality, standardized optical imagery on global scales. For the detection of floating microplastics in the marine environment, however, few studies have succeeded [3]. Previously, limiting factors have included temporal, spatial and spectral coarseness of available data. For example, Landsat 8 provides 9 spectral bands at a spatial resolution of 30 m, with a temporal resolution of 16 days. Commercial satellites including SkySat and RapidEye collect imagery at sub-meter to 5 m spatial resolution, but this is across 3 to 5 spectral bands.

European Space Agency's launched the Sentinel-2A and 2B Earth Observation satellites in 2015 and 2017, respectively, the resolution may now be sufficient to detect floating macroplastics from low-earth orbit. The 12-band Multi-Spectral Instrument (MSI) sensors aboard these satellites were primarily developed for terrestrial services but also cover global coastal waters with revisit times of 2 to 5 days. Their high spatial resolution, up to 10 meters, allows for the detection of small features and objects in the marine environment, such as river plumes, boats, and patches of macroalgae. Moreover, the spatial and spectral resolution of Sentinel-2 enables the differentiation of macroplastics from natural sources of floating debris and seawater [4,5]. The easy access to remote sensing data has stimulated the definition of solution for the detection of marine debris. Many methods are based on optical data and on spectral approach by taking advantage of the multi-spectral nature of the data. In this perspective, Bierman et al. have shown that using simple combination of spectral bands of Sentinel-2 images allows a good discrimination of marine debris. To enhance the detection of marine debris in multispectral satellite data, spectral indices such as the Floating Debris Index (FDI) and the Plastic Index (PI) have been developed, based on artificial plastic targets [6]. Vitale et al. have analyzed how a spectral index based on spectral sign of marine water can make the methods more robust [7]. Following the great interest raised by Sentinel-2 data for marine debris detection, a Marine Debris Archive

(MARIDA) has been recently proposed [13]. MARIDA is a new open-access benchmark dataset based on Sentinel-2 multispectral satellite data. The Marine Debris Archive (MARIDA) is a comprehensive database designed to collect, store, and provide access to data related to marine debris. It incorporates optical imagery from satellites like Sentinel-2, Landsat-8, and commercial satellites, as well as aerial drone data and Integrates data from field studies, including observations from ships, buoys, and underwater vehicles. MARIDA aims to support research, policy-making, and public awareness initiatives by offering a centralized repository of information on marine debris [8]. Although MARIDA has been released specifically for stimulating the development of machine and deep learning solutions for marine debris detection, the aim of this paper is to analyze performance of already existing methods on such datasets. At the moment, all the proposed methods cannot rely on ground truth but only on not precise in-situ indication. MARIDA dataset gives the chance to effectively evaluate the detection ability of proposed method. This work mainly focuses on spectral based ones that will be described in next sections.

The paper is organized as the following: firstly, related works and description of MARIDA dataset are provided in Section II; Section III contains the proposed methodology; Section IV presents the results and discussion; Conclusion are in the last section.

## II. RELATED WORKS

Anthropogenic Marine Debris (AMD) is a major pollutant in the oceans, with millions of tons of debris producing substantial environmental concerns. The growth of marine debris, besides the issues such as heating waters, sea-level rise, and alterations in ocean chemistry, is compromising the condition of marine ecosystem. The detection, study and analysis of marine debris is fundamental for marine ecosystem. In the past, the most of this work has relied on in-situ campaign: samples were collected from beaches on Chiloé Island, Chile, and their spectral signatures and physical properties were analyzed in the lab. This study establishes a novel approach to improve the detection of macroplastics on shoreline using remote sensing techniques [4,9]. These approaches require a huge human effort, time expensive and limited in spatial coverage.

### A. Remote Sensing Approaches

Nowadays, open access to remote sensing data makes airborne and space borne solution a valid alternative for a continuous and rapid monitoring. Moreover, high level performance framework in Python, leveraging libraries such as TensorFlow, Scikit-learn, and GDAL for remote sensing data processing provides a robust framework for developing machine learning solutions for detecting marine floating plastic. The key is to continuously improve the model with new data and validation to ensure high accuracy and reliability in real-world applications [10,11]. A real breakthrough was set by Biermann et al. that in [6] proposed a spectral index-based solution using Sentinel-2 data for debris and plastic detection. In particular, Biermann et al. propose a new spectral index, named Floating Debris Index (FDI), that, being a linear combination of NIR, SWIR and RedEdge bands, is very sensitive to floating material whose spectral signature differs from clean water. This study highlights the feasibility of Sentinel-2 data for plastic detection and the importance of creating a sharing knowledge about data and solutions. To tackle this, a cloud-based framework was developed for extensive marine pollution detection. This framework combines Sentinel-2 satellite imagery with enhanced machine learning tools via the Sentinel Hub cloud application programming interface (API) for detecting ocean plastics at the pilot site on Mytilene Island, Greece [12]. Recently, always considering Sentinel-2 data. Vitale et al. have proposed an advancement of Biermann solution by introducing and additional spectral index based on correlation with water spectral signatures for making prediction more robust. These solutions are more detailed described in Section IV.

### B. Marine Debris Archive (MARIDA)

MARIDA is an open-source dataset that goes in the direction of sharing data and info about marine debris for stimulating research in this field. MARIDA provides real cases of marine debris events with globally distributed annotations, ready for machine learning tasks. This dataset is innovative as it includes various sea features that co-exist in remote sensing images, encompassing a total of 15 thematic classes. Alongside MARIDA, the study presents machine learning baselines for weakly supervised semantic segmentation, featuring both shallow machine learning and deep neural network architectures. Additionally, the multi-label classification task is considered to expand the benchmark's application area. MARIDA can incorporate data from satellites that capture high-resolution images of marine environments. This imagery helps detect large accumulations of marine debris, particularly in remote ocean areas and along coastlines [13,14].

It includes annotated georeferenced polygons and pixels on Sentinel-2 (S2) satellite imagery. Designed for temporal and geographical diversity, MARIDA leverages open-access data from the S2 satellite sensor, which monitors global coastal waters [15]. The S2 satellite can detect and continuously monitor large floating debris, providing multispectral data with spatial resolutions of 10 meters and 20 meters, and a frequent revisit time of 2 to 5 days. For ground-truth events, corresponding S2 level 1C images were obtained from the Copernicus Hub (https://scihub.copernicus.eu/) for the specific reported dates and locations, using an average time window of 10 days. MARIDA can effectively utilize Sentinel-2 data to detect and archive marine floating plastic. This involves acquiring and preprocessing data, extracting relevant features, training a machine learning model, and storing and visualizing the results. Continuous updates and validation of the model with new data will ensure the archive remains accurate and useful for addressing marine debris issues [16,17]. Although MARIDA dataset has been primarily built for the development of ML solutions, this work uses it as a benchmark for assessing the spectral based solution. In particular, the solutions proposed in [6,7] are tested on MARIDA dataset and results are compared with labelled data. The primary objectives of MARIDA are to systematically collect and store data on marine debris from various sources including scientific research, citizen science initiatives, and monitoring programs. It promotes standardized data collection methodologies to ensure consistency and comparability of marine debris data across different studies and regions. It collects data using standardized protocols to ensure data quality and comparability [18,19].

MARIDA is designed to support various remote sensing applications and tasks, with a primary focus on benchmarking weakly supervised pixel-level semantic segmentation learning methods. It provides valuable data for monitoring, analyzing, and mitigating marine debris. Remote sensing technologies, including satellite imagery, aerial surveys, and drone operations, play a crucial role in detecting and tracking marine debris across large and often inaccessible areas. The dataset addresses challenges such as incomplete supervision due to sparsely annotated data, inexact supervision resulting from sensor limitations, and inaccurate supervision caused by potentially slightly noisy annotations [20]. A generic scheme representing the workflow of MARIDA dataset construction is shown in Fig. 1.

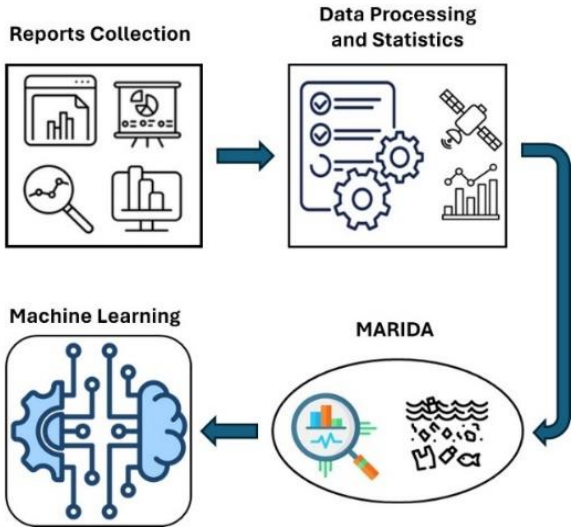

Fig. 1. Chart demonstrating the phases for the formation of Marine Debris Archive (MARIDA).

### III. PROPOSED METHOD

Thanks to MARIDA dataset construction, the aim of this work is to analyze the effectiveness of spectral index-based solutions whose performance have been evaluated only qualitatively, so far. In particular, the performance of two methods is compared: the ones proposed by Biermann et al. in [6] and the one proposed by Vitale et al. in [7]. Both methods rely on the use of spectral index combination for highlighting the presence of floating marine debris. As matter of fact, as previously described in Section II, recently, thanks to availability of multi-spectral remote sensing data, many methods based on spectral characteristics of the data have been proposed. In particular, Biermann et al. [6] explore the potential of the Normalized Difference Vegetation Index (NDVI) and the Floating Debris Index (FDI) on Sentinel-2 data. The NDVI and FDI are derived from combinations of some of the 13 spectral bands of the Sentinel-2 satellite. The idea behind FDI is to design a spectral index that is sensitive to floating materials whose spectral signature differ from clean water. Biermann et al. have demonstrated that by integrating NDVI and FDI allows good detection of marine floating plastic. More precisely, this method aims at detection agglomerate of floating marine debris with the assuming an elevate probability of finding attached plastics.

NDVI is traditionally used to monitor vegetation but can help distinguish vegetation from non-vegetation materials on the water surface. Specifically, the NDVI is defined in Equation 1. Although it is commonly used for highlighting vegetation, this index is also useful for detecting floating agglomerates. As noted by Bierman et al., [6] the NDVI can indicate the presence of agglomerates of mixed materials, such as wood and algae, to which plastic tends to attach.

$$NDVI = \frac{I_8 - I_4}{I_8 + I_4} \qquad (1)$$

Where $I_n$ represents the $n^{th}$ spectral band of Sentinel-2 as shown in the next table.

TABLE I. SPECTRAL BANDS FOR SENTINEL-2 DATA

| *Band* | *Descriptor* | *S2A Wavelength (nm)* | *S2B Wavelength (nm)* | *Resolution (m)* |
|---|---|---|---|---|
| $I_1$ | Coastal | 442.7 | 442.3 | 60 |
| $I_2$ | Blue | 492.4 | 492.1 | 10 |
| $I_3$ | Green | 559.8 | 559.0 | 10 |
| $I_4$ | Red | 664.6 | 665.0 | 10 |
| $I_5$ | Red Edge1 | 704.1 | 703.8 | 20 |
| $I_6$ | Red Edge2 | 740.5 | 739.1 | 20 |
| $I_7$ | Red Edge3 | 782.8 | 779.7 | 20 |
| $I_8$ | NIR | 832.8 | 833.0 | 10 |
| $I_{8a}$ | Narrow NIR | 864.7 | 864.0 | 20 |
| $I_9$ | Water Vapour | 945.1 | 943.2 | 60 |
| $I_{10}$ | SWIR Cirrus | 1373.5 | 1376.9 | 60 |
| $I_{11}$ | SWIR1 | 1613.7 | 1610.4 | 20 |
| $I_{12}$ | SWIR2 | 2202.4 | 2185.7 | 20 |

The FDI has proven to be very useful when combined with the NDVI. The Floating Debris Index (FDI) is an index designed to detect floating debris, such as plastic, on the water surface by utilizing the differences in reflectance between certain spectral bands. The formula for calculating FDI typically uses the red, blue, and near-infrared (NIR) bands.

Given that Sentinel-2 data pixels have resolutions ranging from 10 to 20 meters, it is unlikely for a pixel to be entirely covered by plastic or floating agglomerates. Combining the FDI with the NDVI aids in detecting these materials at the sub-pixel level [21,22,23,24,25].

$$\text{FDI} = I_8 - I'_8 \qquad (2)$$

$$I'_8 = I_6 + (I_{11} - I_6) \times \frac{\lambda_8 - \lambda_4}{\lambda_8 + \lambda_4} \times 10 \qquad (3)$$

Moreover, to reduce the number of false alarms and making the methodology more robust, Vitale et al. [7] propose the additional use of Water Correlation Index (WCI). WCI measures the correlation between spectral sign of marine water and any other object in the scene. The WCI for marine plastic pollution serves as a valuable tool for identifying and quantifying the relationships between plastic pollution levels and various influencing factors. WCI helps in distinguishing water from other materials based on specific band correlations. The Water Correlation Index (WCI) is used to analyze the distribution and concentration of marine plastic pollution. It helps understand how different environmental and anthropogenic factors contribute to the presence and spread of plastic debris in marine environments.

The primary objective of using the Water Correlation Index (WCI) for marine plastic pollution is to understand how different factors contribute to plastic pollution in marine environments. This metric can provide insights into the dynamics of marine plastic pollution and help target interventions more effectively. This index compares the spectral signatures of each pixel, f(p), with that of known seawater, $f_w$. If the correlation exceeds a selected threshold, the pixel is identified as water; otherwise, it is identified as litter.

$$WCI(p) = Corr\,(f(p), f_w) \qquad (4)$$

## IV. RESULTS AND DISCUSSION

So far, the methodology described in the previous section has been validated either with a realistic simulation [21] or using real images characterized by a lack of labelled data. In this section, results of proposed methodology applied to MARIDA dataset are provided.

Two samples from MARIDA dataset have been considered for the testing part and the performance of spectral based methodology have been computed by using either the single spectral indexes or their combination. The two testing cases include clean water (dark blue), man-made structures, like ships, (light blue), wakes (yellow) and marine debris (light yellow). The aim is to show the behavior of FDI, NDVI and WCI on different floating materials.

The first result is shown in Fig. 2: in the first row the RBG image of Sentinel-2 data and labelled mask provided by MARIDA are presented; Results of FDI, NDVI, WCI and their combination are presented in the other rows.

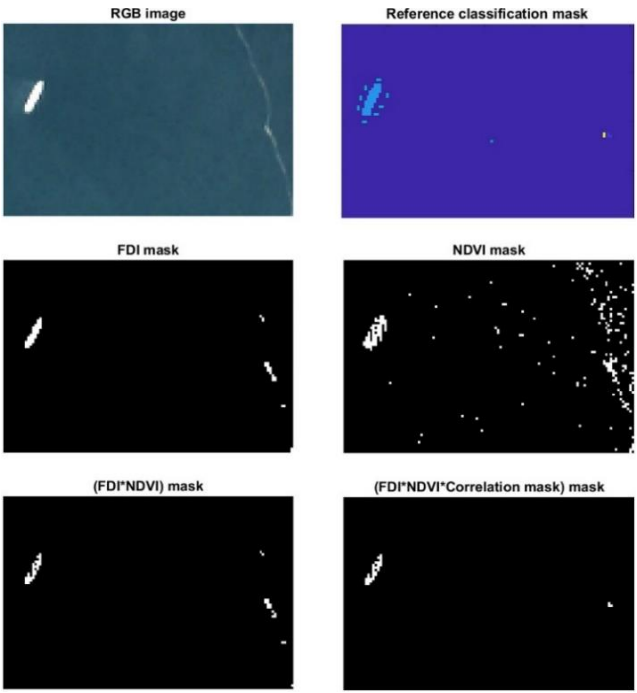

Fig. 2. Test Case 1. RGB image and MARIDA mask in the first row (ship in light blue, marine debris in light yellow). Detection of floating debris with FDI, NDVI on second row. Combination of FDI, NDVI and WCI on last row.

The mask provided by MARIDA highlights the presence of a ship (light blue) and some marine debris (light yellow). In this case the results provided by application of single spectral index are not satisfactory. Indeed, FDI and NDVI produce many false alarms that are corrected by the application of the WCI allowing good discrimination of both the ship and the marine debris. As numerical indicator is considered the accuracy. In this case, the results show an 80% accuracy for the NDVI, 95% for the FDI and 99% for the combination of these two with the WCI.

A second testing case is presented in Fig. 3.

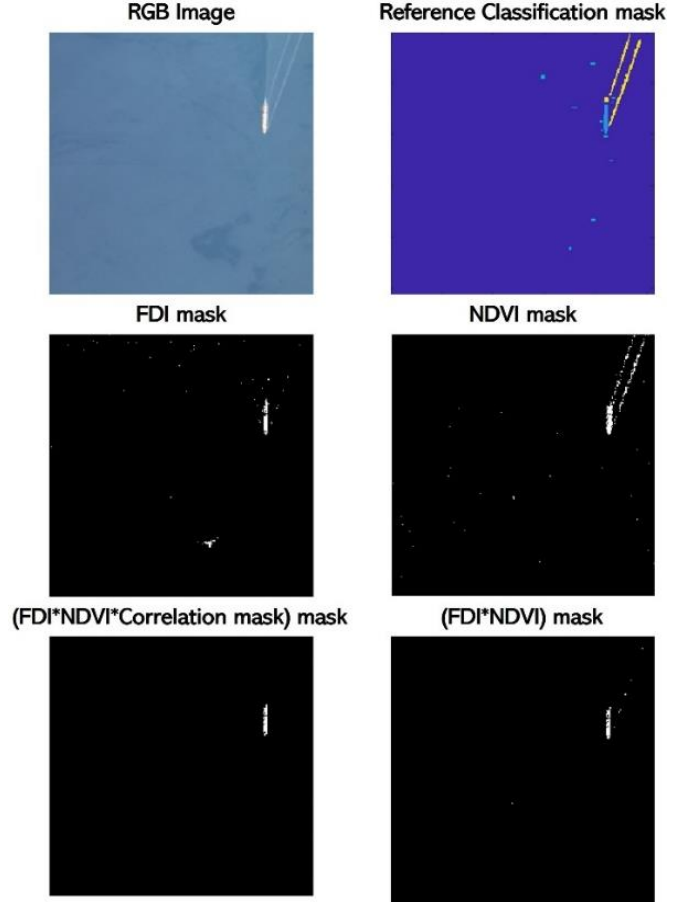

Fig. 3. Test case 2. RGB image and MARIDA mask in the first row (ship in light blue, wakes in yellow). Detection of floating debris with FDI, NDVI on second row. Combination of FDI, NDVI and WCI on last row.

In this case the mask provided by MARIDA highlights the presence of ship (light blue) and of wakes (yellow). It is worthy to note that, in this case, the NDVI looks fundamental for the detection of wakes while FDI and WCI tend to miss such points. As matter of fact, FDI and WCI are designed for detection objects whose spectral sign is different from marine water and the detection of wakes goes beyond their ability as matter of fact, the best accuracy values have been achieved by NDVI with 96%, followed by FDI with 92% and by WCI with 91%.

## V. CONCLUSIONS

The combination of optical remote sensing and ML algorithms offers a powerful approach to detecting and monitoring marine floating plastics. This interdisciplinary approach enhances the ability to understand and address the critical issue of marine pollution, ultimately aiding in the preservation of marine ecosystems. In this work, an assessment of spectral-based solution for marine debris detection is presented. The comparison has been carried out on MARIDA dataset; a benchmark dataset designed for detecting Marine Debris using Sentinel-2 multispectral satellite data. MARIDA poses significant challenges to the research community by offering annotations of Marine Debris alongside various sea features that frequently co-occur in realistic scenarios and providing a comprehensive overview of MARIDA, including spectral signature analyses of the annotated data. For the detection of floating debris, are applied FDI, NDVI and WCI. The results highlight the importance of FDI and WCI for the detection of marine debris but also the difficulty of balancing them for optimal results. Moreover, the presence of a dataset such MARIDA highlights the fundamental importance of having and provided shared data for improving and stimulating research on the preservation of sea environment.